%% file: main.tex
\documentclass[10pt,twocolumn,letterpaper]{article}

\usepackage{cvpr}              
\input{preamble}

\definecolor{cvprblue}{rgb}{0.21,0.49,0.74}
\usepackage[pagebackref,breaklinks,colorlinks,allcolors=cvprblue]{hyperref}

\def\paperID{39} 
\def\confName{CVPR}
\def\confYear{2026}

\title{Does Your VFM Speak Plant? The Botanical Grammar of Vision Foundation Models for Object Detection}

\author{Lars Lundqvist, Earl Ranario, Hamid Kamangir, Heesup Yun, Christine Diepenbrock, \and Brian N. Bailey, J. Mason Earles\\
University of California, Davis\\
{\tt\small \{llund, ewranario, hkamangir, hspyun, chdiepenbrock, bnbailey, jmearles\}@ucdavis.edu}
}

\begin{document}
\maketitle
\input{sec/0_abstract}
\input{sec/updated_core}
{
    \small
    \bibliographystyle{ieeenat_fullname}
    \bibliography{main}
}


\end{document}

%% file: preamble.tex
\makeatletter
\@ifpackageloaded{lineno}{\switchlinenumbers}{}
\makeatother

\usepackage{placeins}

%% file: sec/0_abstract.tex
\begin{abstract}
    Vision foundation models (VFMs) offer the promise of zero-shot object detection without task-specific training data, yet their performance in complex agricultural scenes remains highly sensitive to text prompt construction. We present a systematic prompt optimization framework evaluating four open-vocabulary detectors---YOLO World, SAM3, Grounding DINO, and OWLv2---for cowpea flower and pod detection across synthetic and real field imagery. We decompose prompts into eight axes and conduct one-factor-at-a-time analysis followed by combinatorial optimization, revealing that models respond divergently to prompt structure: conditions that optimize one architecture can collapse another. Applying model-specific combinatorial prompts yields substantial gains over a na\"ive species-name baseline, including $+0.357$ mAP@0.5 for YOLO World and $+0.362$ mAP@0.5 for OWLv2 on synthetic cowpea flower data. To evaluate cross-task generalization, we use an LLM to translate the discovered axis structure to a morphologically distinct target---cowpea pods---and compare against prompting using the discovered optimal structures from synthetic flower data. Crucially, prompt structures optimized exclusively on synthetic data transfer effectively to real-world fields: synthetic-pipeline prompts match or exceed those discovered on labeled real data for the majority of model--object combinations (flower: 0.374 vs.\ 0.353 for YOLO World; pod: 0.429 vs.\ 0.371 for SAM3). Our findings demonstrate that prompt engineering can substantially close the gap between zero-shot VFMs and supervised detectors without requiring manual annotation, and that optimal prompts are model-specific, non-obvious, and transferable across domains.
\end{abstract}

%% file: sec/updated_core.tex
\section{Introduction}
\label{sec:intro}

Modern vision foundation models (VFMs) claim to enable strong zero-shot performance on various unseen classes. These models are pretrained on expansive datasets covering many common scenes and trained to enable out-of-distribution inference performance, acting as open-vocabulary (or open-set) detectors (OVDs). However, agricultural scenes present a highly complex scenario for testing of this model generalization~\cite{AlNahian2025AgriFM}. The objects of interest (\eg flowers, pods, leaves) are typically small, visually similar to surrounding plant structures, and ideal detection performance can vary significantly based on the user's class definitions (\eg single-class flowers versus flowers at different growth stages). The current question to answer is how these VFMs can be effectively applied in this scenario while minimizing end-user intervention. To provide true zero-shot usage of these models, methods must be found for optimization of OVD performance without required labeled data for finetuning. This problem will become increasingly relevant as the zero-shot performance of VFMs continues to improve, and questions remain on the ideal translation between an expert's description of the class of interest and what will best be ``understood'' by a model.

We approach the problem by first aiming to understand which ``standard'' prompt would produce the best performance for four VFMs: YOLO World (yolov8x-worldv2)~\cite{Cheng2024YOLOWorld}, SAM3~\cite{carion2025sam3segmentconcepts}, Grounding DINO (grounding-dino-base)~\cite{Liu2023GroundingDINO}, and OWLv2 (owlv2-base-patch16-ensemble)~\cite{Minderer2023OWLv2}. We decompose prompts into interpretable linguistic axes (taxonomy, color, anatomy, phenology, grammar, size, negation) and one non-linguistic axis (emoji). We include emoji based on evidence that they influence text-conditioned vision models~\cite{Wen2023HardPromptsMadeEasy}, plausibly because emoji are prevalent in the large web-scraped caption datasets used to train text encoders such as CLIP's~\cite{radford2021learningtransferablevisualmodels}. We then conduct a systematic one-factor-at-a-time analysis followed by combinatorial optimization of the best-performing components. From this, we observe a divergence in ideal prompting strategies between models. Notably, our benchmarking of ideal prompt structuring holds for both synthetic and real cowpea datasets, suggesting that the discovered prompt structures capture generalizable information about how these models encode agricultural concepts. We further validate this generalization by applying the pipeline to cowpea pod detection via LLM-guided axis translation, demonstrating that prompts optimized entirely on synthetic data match or exceed those discovered by running the full pipeline on labeled real data.
\subsection{Related Work}
\label{sec:related_work}

\subsubsection{Zero and Few-Shot Detection in Agriculture}
Singh~\etal~\cite{Singh2025FewShotGroundingDINOAgri} investigated few-shot adaptation of Grounding DINO for agricultural object detection by removing the text input, instead initializing the text embedding randomly, and conditioning on a small set of labeled input images. Their zero-shot experiments explicitly revealed challenges in crafting effective text prompts for complex agricultural objects such as individual leaves and visually similar classes, motivating their shift to few-shot adaptation. While this method showed strong performance with few labeled examples, it still requires some labeled data, which our work aims to avoid entirely through prompt engineering. Mullins~\etal~\cite{Mullins2024ZeroShotBlueberry} demonstrated the difficulties of using these models in zero-shot settings with fixed prompts for wild blueberries, finding ideals of ``small blue sphere'' and ``smooth blueberry'' for YOLO World~\cite{Cheng2024YOLOWorld} and Grounding DINO~\cite{Liu2023GroundingDINO} respectively. Their finding that optimal prompts differ substantially between models and diverge from standard class descriptions directly motivates the systematic axis decomposition we pursue in this work.

\subsubsection{Prompt Engineering for Vision-Language Models}
Prompt engineering for VLMs has been studied extensively in the general computer vision literature. Gu~\etal~\cite{Gu2023PromptSurveyVLM} provide a systematic survey covering hard prompts (task instructions, in-context learning, chain-of-thought) and soft prompts (prompt tuning, prefix token tuning) across multimodal-to-text, image-text matching, and text-to-image models. A key distinction in this is between hard prompts (discrete natural language strings that can be manually crafted or searched) and soft prompts (continuous vector representations optimized via gradient descent). Our work operates entirely in the hard prompt regime, requiring no gradient access to the target model.

On the soft-prompting side, Zhou~\etal~\cite{Zhou2022CoOp} introduced Context Optimization (CoOp), which replaces context words in CLIP prompts with learnable vectors and achieves large improvements over hand-crafted prompts with as few as one or two labeled shots. They note that prompt engineering for VLMs requires significant effort from small changes in semantics producing large increases in performance in poorly understood ways, a challenge our systematic axis framework directly addresses. Zhou~\etal~\cite{Zhou2022CoCoOp} subsequently proposed Conditional Context Optimization (CoCoOp), which generates input-conditional prompt tokens to improve generalization to unseen classes. Our approach is complementary: the structured hard prompts we discover could serve as informed initialization points for soft-prompt methods, and the axis decomposition itself requires no labeled data whatsoever.

On the hard prompt optimization side, Mahajan~\etal~\cite{Mahajan2023PromptingHardOrHardlyPrompting} studied prompt inversion for text-to-image diffusion models, revealing the abstract and often human-uninterpretable nature of prompts that produce optimal outputs. Wen~\etal~\cite{Wen2023HardPromptsMadeEasy} further demonstrated tractable discrete optimization in text-conditioned vision pipelines. Leviathan~\etal~\cite{Leviathan2025PromptRepetition} found that non-reasoning LLMs saw performance improvements from prompt repetition, but did not explore this phenomenon in vision tasks. These findings on prompt sensitivity and optimization in adjacent domains motivate our systematic investigation of how prompt structure affects zero-shot detection in agriculture.

\subsubsection{Prompt Optimization versus Finetuning}
A central motivation of this work is the practical accessibility of prompt-based adaptation compared to model finetuning. Finetuning vision-language models, whether through full parameter updates, LoRA~\cite{Hu2022LoRA}, or other parameter-efficient methods, requires labeled training data, GPU compute for backpropagation, and ML engineering expertise that may be unavailable to agricultural end-users. Even parameter-efficient approaches like LoRA, which have been applied to agricultural vision tasks~\cite{ESPEJOGARCIA2025110900}, require at minimum a curated labeled dataset and a training loop. By contrast, prompt engineering operates at inference time with no parameter updates, making it immediately deployable by domain experts who can describe their target objects in natural language but lack ML infrastructure. Our pipeline further reduces the expertise barrier by automating the prompt search through structured axis decomposition and LLM-guided translation, requiring only a target class description as input.

\begin{table*}[t]
  \caption{Systematic prompt axis definitions for one-factor-at-a-time (OFAT) analysis for cowpea flowers. Baseline values represent the control group used as a fixed anchor during single-axis perturbations. The final prompt is constructed as: \textit{[Grammar] [Color] [Taxonomy] [Anatomy] [Phenology] [Negation] [Emoji]}.}
  \label{tab:ofat_axes}
  \centering
  \small
  \begin{tabular*}{\textwidth}{@{\extracolsep{\fill}}lllp{10cm}@{}}
\toprule
\textbf{Axis} & \textbf{Levels} & \textbf{Baseline} & \textbf{Perturbation Values} \\
\midrule
Taxonomy  & 8 & flower & \{cowpea, bean, pea, legume, black-eyed pea, vigna unguiculata, crop\} flower \\
\addlinespace
Color     & 5 & none   & yellow, white, cream, purple \\
\addlinespace
Size      & 4 & none   & large, small, tiny \\
\addlinespace
Phenology & 6 & none   & bud, open, closed bud, blooming, in bloom \\
\addlinespace
Negation  & 6 & none   & full negation (all confusers), partial negation combinations \\
\addlinespace
Anatomy   & 5 & none   & with open petals, with visible petals, with petals and stamens, corolla \\
\addlinespace
Grammar   & 6 & a      & a single, bare noun, a photo of a, one, close-up of a \\
\addlinespace
Emoji     & 6 & none   & cherry blossom, hibiscus, sunflower, blossom, bouquet, tulip \\
\bottomrule
  \end{tabular*}
\end{table*}

\begin{table*}[t]
\centering
\caption{Phase~2 combinatorial generation strategy. Base combinations are generated programmatically from the three sweeps below; negation and emoji variants are then appended to the top-ranked results. Axes marked with $\star$ use the best value identified in Phase~1 OFAT analysis.}
\label{tab:combination_strategy}
\small
\begin{tabular*}{\textwidth}{@{\extracolsep{\fill}}llll}
\toprule
\textbf{Sweep} & \textbf{Varied Axes} & \textbf{Fixed Axes} & \textbf{Prompt Template} \\
\midrule
1.\ Color $\times$ Size & Color, Size & Grammar$^\star$, Taxonomy$^\star$ & [gram] [size] [color] [taxonomy] \\
\addlinespace
2.\ Grammar $\times$ Color & Grammar, Color & Taxonomy$^\star$ & [gram] [color] [taxonomy] \\
\addlinespace
3.\ Anatomy & Anatomy & Grammar$^\star$, Color$^\star$, Taxonomy$^\star$ & [gram] [color] [taxonomy] [anatomy] \\
\midrule
\addlinespace
Negation (top-$N$) & Negation & Best base prompt & [base prompt], [negation] \\
\addlinespace
Emoji (top-1) & Emoji & Best prompt (incl.\ negation) & [best prompt] [emoji] \\
\bottomrule
\end{tabular*}
\end{table*}

\section{Method}
\label{sec:method}

\subsection{Phase 1: Factor Analysis}
\label{sec:phase1}

Based on the poor, highly-varying zero-shot performance observed for prompts matching the desired labels (\eg ``a flower''), we introduced a multi-phase prompt-testing suite that varies the prompt along axes relevant to the detection task. In the first phase, we conduct a one-factor-at-a-time (OFAT) analysis across eight prompt component axes to isolate the contribution of each component per model architecture. The axes and their perturbation values are shown in Table~\ref{tab:ofat_axes}. The baseline prompt is ``a flower'' with all axes at their default values. This yields 40 configurations per model (1 baseline + 39 axis-level variations), allowing us to identify which components matter most for each architecture and how prompts should be structured.

\subsection{Phase 2: Combinatorial Optimization}
\label{sec:phase2}

In the second phase, we programmatically generate combinatorial prompt configurations from the Phase~1 OFAT results and evaluate them to find prompts better suited to each model's preferences. The generation procedure first identifies the best-performing grammar and taxonomy values from the Phase~1 scores (achieving the highest mAP@0.5 among OFAT combinations varying only that axis). These serve as anchors for three systematic sweeps:

\begin{enumerate}
    \item \textbf{Color $\times$ Size}: all non-baseline color and size values are crossed at the best taxonomy and best grammar, producing prompts of the form \textit{[grammar] [size] [color] [taxonomy]}.
    \item \textbf{Grammar $\times$ Color}: all grammar values (excluding the Phase~1 best) are crossed with all non-baseline colors at the best taxonomy, testing whether alternative grammars interact favorably with color descriptors.
    \item \textbf{Anatomy variants}: each anatomy value is combined with the best color, grammar, and taxonomy, producing prompts of the form \textit{[grammar] [color] [taxonomy] [anatomy]}.
\end{enumerate}

After evaluating all base combinations, the pipeline ranks results by mAP@0.5 and appends negation variants to the top-$N$ prompts (default $N=3$). Each negation value from the axis is concatenated to the base prompt (\eg ``\dots, not a leaf, not a stem''), and the resulting prompts are evaluated independently. Finally, emoji variants are generated for the single top-scoring prompt (across both base and negation results) by appending each emoji token from the axis. This staged approach avoids a combinatorial explosion while ensuring that negation and emoji---which interact strongly with prompt content---are tested against the most promising candidates rather than exhaustively. Table~\ref{tab:combination_strategy} summarizes the generation strategy.

\subsubsection{Background Class Regularization}
A consistent finding across all YOLO World experiments is that appending an
empty string (\texttt{""}) as an additional background class to the prompt
list significantly improves detection performance. This background absorber
class acts as a catch-all that competes with the target prompt(s) for
low-confidence detections, effectively raising the decision boundary and
suppressing spurious false positives. We include this empty-string class in
all YOLO World configurations reported in this work.
Without this regularization, YOLO World produces substantially more false
positives, particularly on visually cluttered agricultural scenes where plant
structures such as calyces and leaf edges generate weak but nonzero
confidence scores against the target prompt.

\subsubsection{Automated Axis Translation via LLM}

To evaluate the generalizability of our benchmarking strategy across different agricultural tasks, we developed an automated pipeline that leverages a large language model (LLM) to translate the baseline linguistic axes to novel target classes. We utilize Qwen-4B to translate the original \textit{cowpea flower} factor axes into domain-appropriate values for new objects, such as \textit{cowpea pods}.

This process requires only the new target crop or object as an input. The LLM preserves the structural axes (taxonomy, color, size, phenology, negation, anatomy, grammar, and emoji) while systematically updating the prompt values and baselines to match the new target. For example, negation axes are dynamically populated with the most likely confuser classes for the new object. Phase 1 (OFAT) and Phase 2 (Combinatorial) testing are then executed programmatically on these translated axes. This allows for rapid, zero-shot scaling of the optimization strategy to new crops and structures without manual prompt engineering.

\subsection{Dataset and Metrics}
\label{sec:dataset}
For flower experiments, we run the framework on a synthetic dataset of 500 cowpea images containing 3783 flowers generated using the Helios plant simulation software~\cite{Helios, LEI20240189, bailey2025generalizedframeworkproceduralgeneration}, and a dataset of 268 real cowpea images containing 1826 annotated flowers. For pod generalization experiments, we additionally run the two-phase benchmark on a Helios-generated synthetic dataset of 499 images containing 5066 pods and a dataset of 186 real cowpea images containing 1603 pods. All zero-shot results are compared against supervised YOLOv11~\cite{Jocher2024YOLOv11} baselines trained on the same real-world experiments, which achieve mAP@0.5 of 0.763 and 0.651 on the real flower and pod datasets, respectively. COCO-style mAP~\cite{Lin2014COCO} is used throughout benchmarking to determine the optimal prompts to use while avoiding issues related to confidence threshold selection. We report mAP@0.5 rather than mAP@0.5:0.95 because accurate detection counts matter more than precise bounding box placement in this application context.

\begin{figure*}[t!]
    \centering
    \includegraphics[width=\textwidth]{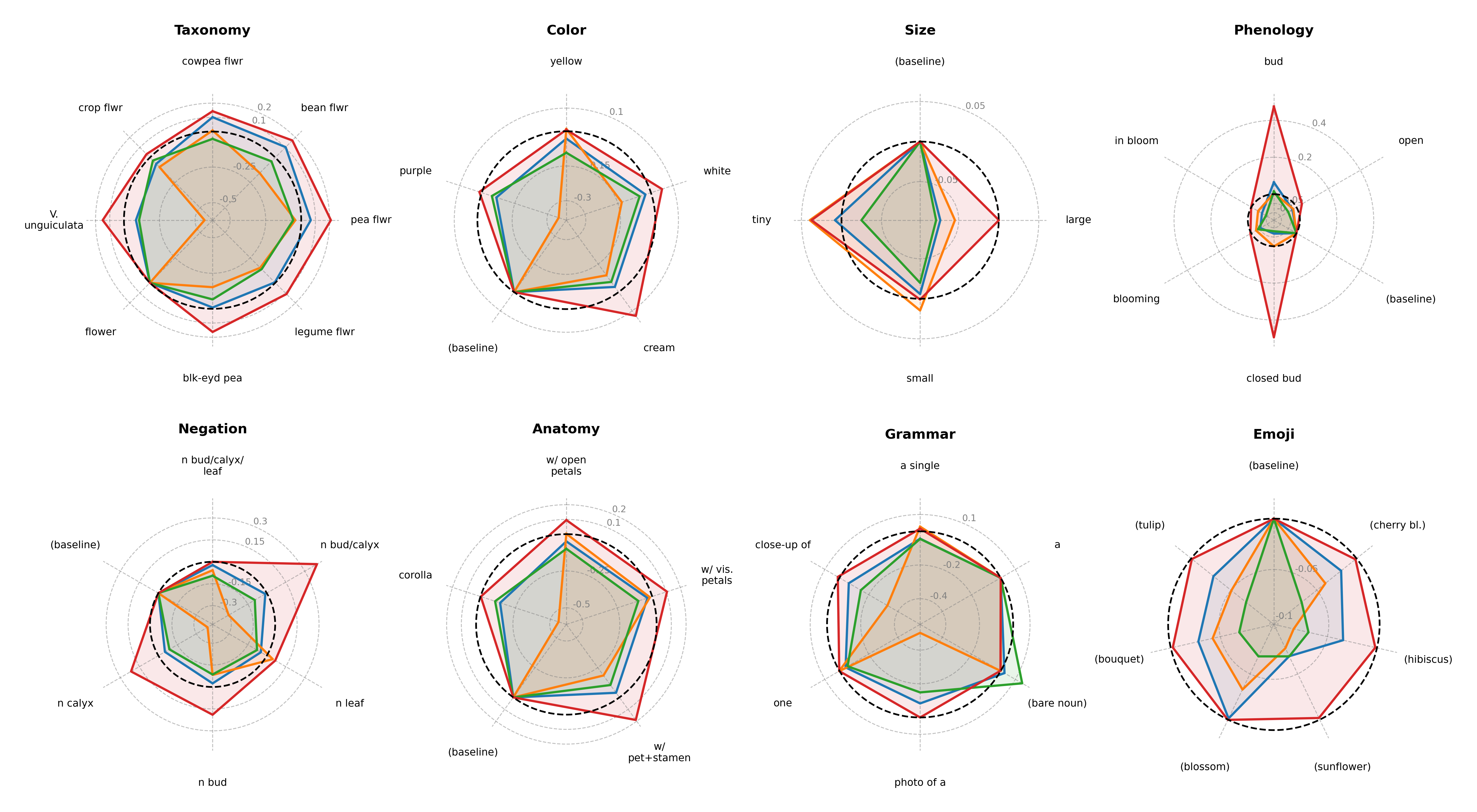}
    \caption{Phase~1 OFAT analysis: change in mAP@0.5 relative to
    each model's baseline prompt (dashed circle at $\Delta = 0$).
    Each plot shows one prompt axis; vertices represent
    perturbation values from Table~\ref{tab:ofat_axes}.
    Legend: \textcolor[HTML]{4472C4}{YOLO World},
    \textcolor[HTML]{ED7D31}{SAM3},
    \textcolor[HTML]{548235}{Grounding DINO},
    \textcolor[HTML]{FF0000}{OWLv2}.
    }
    \label{fig:ofat-radar}
\end{figure*}

\section{Experiments}
\label{sec:experiments}
\begin{table*}[t]
\centering
\caption{Phase 1 Benchmark Results: Best Single-Prompt Performance. $\Delta$ represents the mAP@0.5 improvement over the generic baseline prompt (``a flower''). \textit{Baseline YOLOv11 achieves $0.763$ mAP@0.5 on real}.}
\label{tab:benchmarking_results}
\small
\begin{tabular*}{\textwidth}{@{\extracolsep{\fill}}lllll}
\toprule
\textbf{Model} & \textbf{Dataset} & \textbf{Best Single Prompt} & \textbf{mAP@0.5} & \textbf{$\Delta$ (Gain)} \\
\midrule
SAM3 & Synthetic & ``a flower not a leaf'' & 0.609 & +0.072 \\
OWLv2 & Synthetic & ``a closed bud'' & 0.471 & +0.469 \\
Grounding DINO & Synthetic & ``flower'' & 0.288 & +0.137 \\
YOLO World & Synthetic & ``a bean flower'' & 0.222 & +0.081 \\
\addlinespace
\midrule
SAM3 & Real & ``a closed bud'' & 0.491 & +0.123 \\
OWLv2 & Real & ``a closed bud'' & 0.350 & +0.349 \\
Grounding DINO & Real & ``flower'' & 0.169 & +0.128 \\
YOLO World & Real & ``a cowpea flower'' & 0.151 & +0.090 \\
\bottomrule
\end{tabular*}
\end{table*}

\subsection{Phase 1: Single-Axis Factor Analysis}
\label{sec:exp_phase1}

The OFAT analysis reveals that models respond to prompt axes in markedly different ways. For YOLO World and SAM3, the taxonomy and color axes produce the largest improvements, while Grounding DINO is largely insensitive to prompt elaboration beyond the bare noun ``flower.'' OWLv2 shows an unusual sensitivity to phenology terms, with ``a closed bud'' dramatically outperforming other single-axis prompts. We hypothesize that this result reflects a scale bias in pretraining data: buds are typically small relative to the frame, aligning with the small apparent size of cowpea flowers in field imagery. However, prompting for ``bud'' in a deployment context would introduce false positives from undeveloped flowers and the abundant small green structures (calyces, leaf tips) present in these scenes, making it unsuitable as a practical detection strategy despite its high mAP in isolation. These divergent responses underscore that no single prompting strategy is universally optimal and motivate the per-model combinatorial optimization in Phase~2.

Table~\ref{tab:benchmarking_results} summarizes the best single-prompt performance for each model on both datasets. The $\Delta$ column reports the mAP improvement relative to the generic baseline prompt ``a flower.'' Several findings are notable. First, the best-performing prompt differs not only between models but also between datasets for the same model: SAM3 achieves its best real-world performance with ``a closed bud'' rather than the ``a flower not a leaf'' that dominates on synthetic data. Second, OWLv2 achieves near-zero mAP on real data with the baseline prompt but jumps to 0.348 with ``a closed bud,'' demonstrating extreme prompt sensitivity. Third, Grounding DINO shows the least responsiveness to prompt variation, with the bare noun ``flower'' remaining optimal across both datasets.

\subsection{Phase 2: Combinatorial Prompt Optimization}
\label{sec:exp_phase2}
\begin{table*}[t]
\centering
\caption{Phase~2 best combinatorial prompts. $\Delta$: mAP@0.5 gain over ``a flower'' baseline. Emoji rows (\dag) show the best prompt with an appended emoji token.}
\label{tab:phase2_results}
\small
\begin{tabular*}{\textwidth}{@{\extracolsep{\fill}}llcccc}
\toprule
 & & \multicolumn{2}{c}{\textbf{Synthetic}} & \multicolumn{2}{c}{\textbf{Real}} \\
\cmidrule(lr){3-4} \cmidrule(lr){5-6}
\textbf{Model} & \textbf{Best Prompt} & \textbf{mAP} & \textbf{$\Delta$} & \textbf{mAP} & \textbf{$\Delta$} \\
\midrule
YOLO World & a single yellow cowpea flower w/ open petals, not bud, calyx, leaf & 0.433 & +0.292 & 0.299 & +0.238 \\
\hspace{1em}\dag\,+ (bouquet) & & 0.498 & +0.357 & 0.374 & +0.313 \\
\addlinespace
SAM3 & a single yellow cowpea flower w/ visible petals & 0.635 & +0.098 & 0.532 & +0.164 \\
\hspace{1em}\dag\,+ (blossom) & & 0.649 & +0.112 & 0.544 & +0.176 \\
\addlinespace
OWLv2 & a single yellow pea flower with open petals, not a bud & 0.364 & +0.362 & 0.325 & +0.324 \\
\addlinespace
Grounding DINO & small flower & 0.208 & +0.057 & 0.085 & +0.044 \\
\bottomrule
\multicolumn{6}{@{}p{\textwidth}}{\footnotesize\dag\,For OWLv2 and Grounding DINO, emoji combinations did not outscore non-emoji configurations in Phase~2 synthetic benchmarking and were therefore not evaluated on real data.} \\
\end{tabular*}
\end{table*}

Combining top-performing axes yields substantial gains for most models. Table~\ref{tab:phase2_results} reports the best combinatorial prompt for each model on the synthetic dataset, with $\Delta$ computed relative to the baseline ``a flower''.

YOLO World benefits most from prompts with full text negation, achieving an mAP of 0.433 (+0.292 over baseline). The inclusion of negation clauses (``not a bud, not the green calyx, not a leaf'') proves critical for suppressing false positives from visually similar plant structures. SAM3 and OWLv2 both achieve their best performance from similar prompts utilizing ``a single'' as their grammar and describing color and anatomy, but show less need for full negation. Grounding DINO remains unresponsive to added information, with the best Phase 2 prompt of ``small flower'' still underperforming ``flower'' from Phase 1.

The inclusion of emoji characters produces further gains for YOLO World and
SAM3. When flower-related emoji are appended to the combinatorial prompts,
SAM3 reaches 0.649 mAP (+0.112 over ``a flower'') and YOLO World
achieves 0.498 (+0.357) on synthetic data. For OWLv2 and Grounding DINO,
emoji-augmented variants did not outscore their non-emoji counterparts in
synthetic benchmarking and were not carried forward to real-data evaluation.

Crucially, though these tests are performed using the synthetic dataset only, the performance improvement of discovered prompt configurations transfers to the unseen real-world dataset, suggesting the discovered prompt structures capture generalizable information about how these models encode the target class rather than dataset-specific information.

\subsubsection{Generalization to Cowpea Pods}
\label{sec:exp_pod_generalization}

To evaluate whether the structural prompt preferences discovered for flowers generalize to a different agricultural object, we apply the pipeline to cowpea pod detection on real field imagery. We compare three prompting strategies: (i) a na\"ive baseline (``a pod''), (ii) structured prompts assembled by manually translating the best-performing flower axis configurations from Phase~1 into pod-appropriate terms, and (iii) prompts discovered by the full LLM-guided synthetic pipeline (Phases~1 and~2 executed on LLM-translated axes). All results are evaluated on real pod imagery against a supervised YOLOv11 baseline (mAP@0.5 = 0.651). Table~\ref{tab:pod_results} reports the results.

The structured prompts, constructed by preserving the optimal axis structure for each model and substituting flower-specific terms with pod-specific ones, follow directly from the flower benchmarking. For YOLO World, where negation was the dominant factor, the structured prompt is ``a single green cowpea pod with peas, not a flower, not a leaf, not a stem.'' For SAM3, which favored the anatomy descriptor pattern, the structured prompt is ``a single green cowpea pod with visible peas.'' OWLv2 uses ``a single green pea pod with peas, not a stem,'' and Grounding DINO uses ``pod,'' reflecting its preference for minimal prompts.

The LLM-generated prompts from the full synthetic pipeline yield the strongest performance for three of four models (Table~\ref{tab:pod_results}). SAM3 shows the most striking gain: the LLM-translated axes introduced the color descriptor ``mottled'', producing the prompt ``a single mottled pea pod, not a leaf, not a stem,'' which achieves 0.429 mAP---a substantial improvement over both the structured prompt (0.100) and the baseline (0.039). This result highlights the value of LLM-guided axis population: the term ``mottled'' is unlikely to appear in a manual translation from flower axes, yet it seems to capture a relevant description of pods from the training set that SAM3's text encoder can exploit. For YOLO World, the synthetic pipeline prompt (0.050) also outperforms the na\"ive baseline (0.000) but falls short of the manually structured prompt (0.065), suggesting that for some model--object combinations, domain expertise in prompt assembly can still complement automated discovery.

\begin{table*}[t]
\centering
\caption{Pod detection results on real data (mAP@0.5). Three prompting strategies are compared: na\"ive baseline, structured prompts translated from flower Phase~1 axis preferences, and prompts discovered by the full LLM-guided synthetic pipeline. Best zero-shot result per model in \textbf{bold}. \textit{Baseline YOLOv11: mAP@0.5} $= 0.651$}
\label{tab:pod_results}
\small
\begin{tabular*}{\textwidth}{@{\extracolsep{\fill}}lccc}
\toprule
\textbf{Model} & \textbf{Baseline (``a pod'')} & \textbf{Structured Prompt} & \textbf{Best Synthetic Prompt} \\
\midrule
YOLO World      & 0.000 & \textbf{0.065} & 0.050 \\
SAM3            & 0.039 & 0.100 & \textbf{0.429} \\
OWLv2           & 0.002 & 0.071 & \textbf{0.111} \\
Grounding DINO  & 0.007 & 0.010 & \textbf{0.036} \\
\bottomrule
\end{tabular*}
\end{table*}

\subsubsection{Synthetic-to-Real Prompt Transfer}
\label{sec:exp_syn_real_transfer}

A key practical question is whether running the full optimization pipeline on synthetic data alone produces prompts comparable to those discovered by running the same pipeline directly on real labeled data. To quantify this generalization gap, we execute the complete Phase~1 and Phase~2 pipeline on the real datasets for both flowers and pods, and compare the resulting best prompts against those discovered synthetically (Table~\ref{tab:syn_real_transfer}).

For flowers, synthetic-discovered prompts match or exceed real-discovered prompts for three of four models: YOLO World achieves 0.374 (synthetic) versus 0.353 (real), SAM3 achieves 0.544 versus 0.554, and OWLv2 achieves 0.325 versus 0.402. Grounding DINO shows a larger gap (0.085 synthetic versus 0.049 real), though both values remain low, consistent with this model's general insensitivity to prompt optimization. The near-parity for YOLO World and SAM3 confirms that the prompt structures optimized on synthetic imagery capture model-level encoding preferences rather than dataset-specific artifacts.

For pods, the pattern is similar: SAM3's synthetic prompt substantially outperforms the real-discovered prompt (0.429 versus 0.371), driven by the LLM-introduced ``mottled'' descriptor. OWLv2 likewise favors the synthetic prompt (0.111 versus 0.085). YOLO World shows a modest advantage for the real-discovered prompt (0.094 versus 0.050), and Grounding DINO performs poorly under both strategies. Across both object classes and all four models, the synthetic pipeline produces competitive or superior prompts in the majority of cases, supporting the practical viability of prompt optimization without any labeled real-world data.

\begin{table*}[t]
\centering
\caption{Synthetic-to-real prompt transfer comparison (mAP@0.5 on real data). ``Syn.\ Pipeline'' uses prompts discovered by running Phases~1--2 on synthetic data; ``Real Pipeline'' uses prompts from the same pipeline executed on labeled real data. Best result per model--object pair in \textbf{bold}.}
\label{tab:syn_real_transfer}
\small
\begin{tabular*}{\textwidth}{@{\extracolsep{\fill}}lcccc}
\toprule
 & \multicolumn{2}{c}{\textbf{Flowers}} & \multicolumn{2}{c}{\textbf{Pods}} \\
\cmidrule(lr){2-3} \cmidrule(lr){4-5}
\textbf{Model} & \textbf{Syn.\ Pipeline} & \textbf{Real Pipeline} & \textbf{Syn.\ Pipeline} & \textbf{Real Pipeline} \\
\midrule
YOLO World      & \textbf{0.374} & 0.353 & 0.050 & \textbf{0.094} \\
SAM3            & 0.544 & \textbf{0.554} & \textbf{0.429} & 0.371 \\
OWLv2           & 0.325 & \textbf{0.402} & \textbf{0.111} & 0.085 \\
Grounding DINO  & \textbf{0.085} & 0.049 & \textbf{0.036} & 0.013 \\
\bottomrule
\end{tabular*}
\end{table*}

\begin{figure*}[t]
    \centering
    \setlength{\tabcolsep}{4pt}
    \renewcommand{\arraystretch}{0.6}
    \begin{tabular}{@{}cc@{}}
        \small\textbf{YOLO World} & \small\textbf{SAM3} \\[3pt]
        \includegraphics[width=0.47\textwidth]{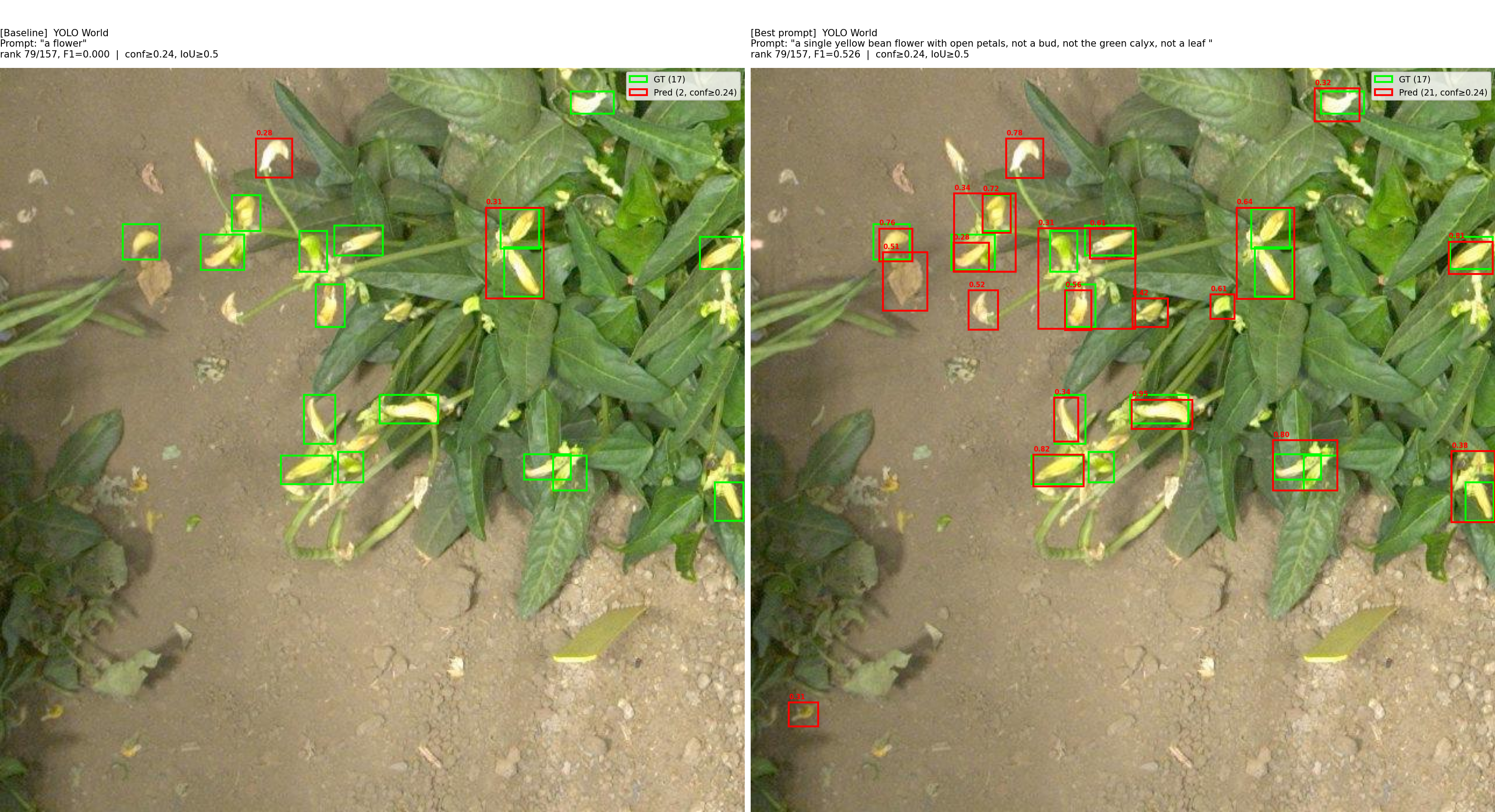} &
        \includegraphics[width=0.47\textwidth]{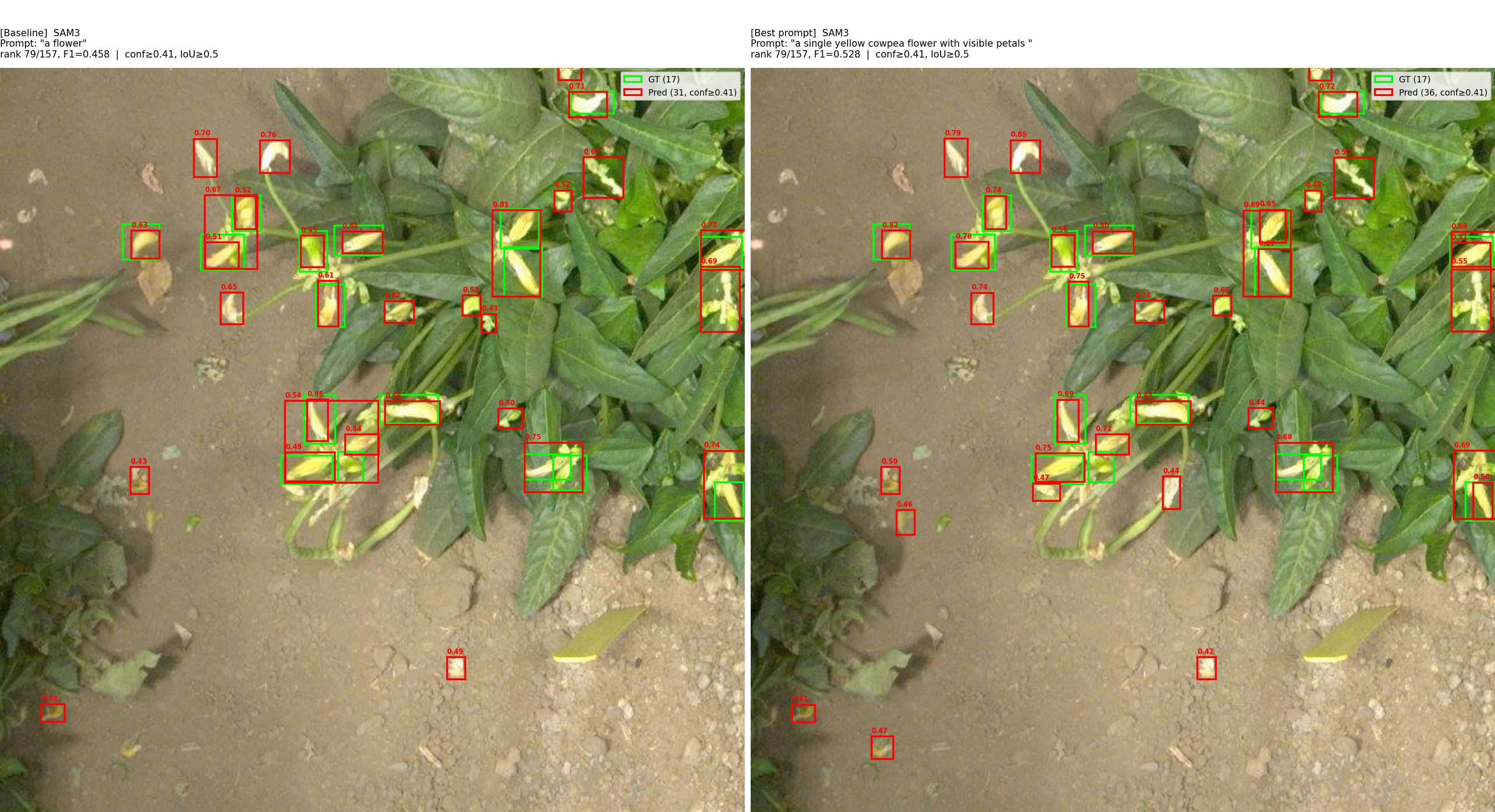} \\[6pt]
        \small\textbf{OWLv2} & \small\textbf{Grounding DINO} \\[3pt]
        \includegraphics[width=0.47\textwidth]{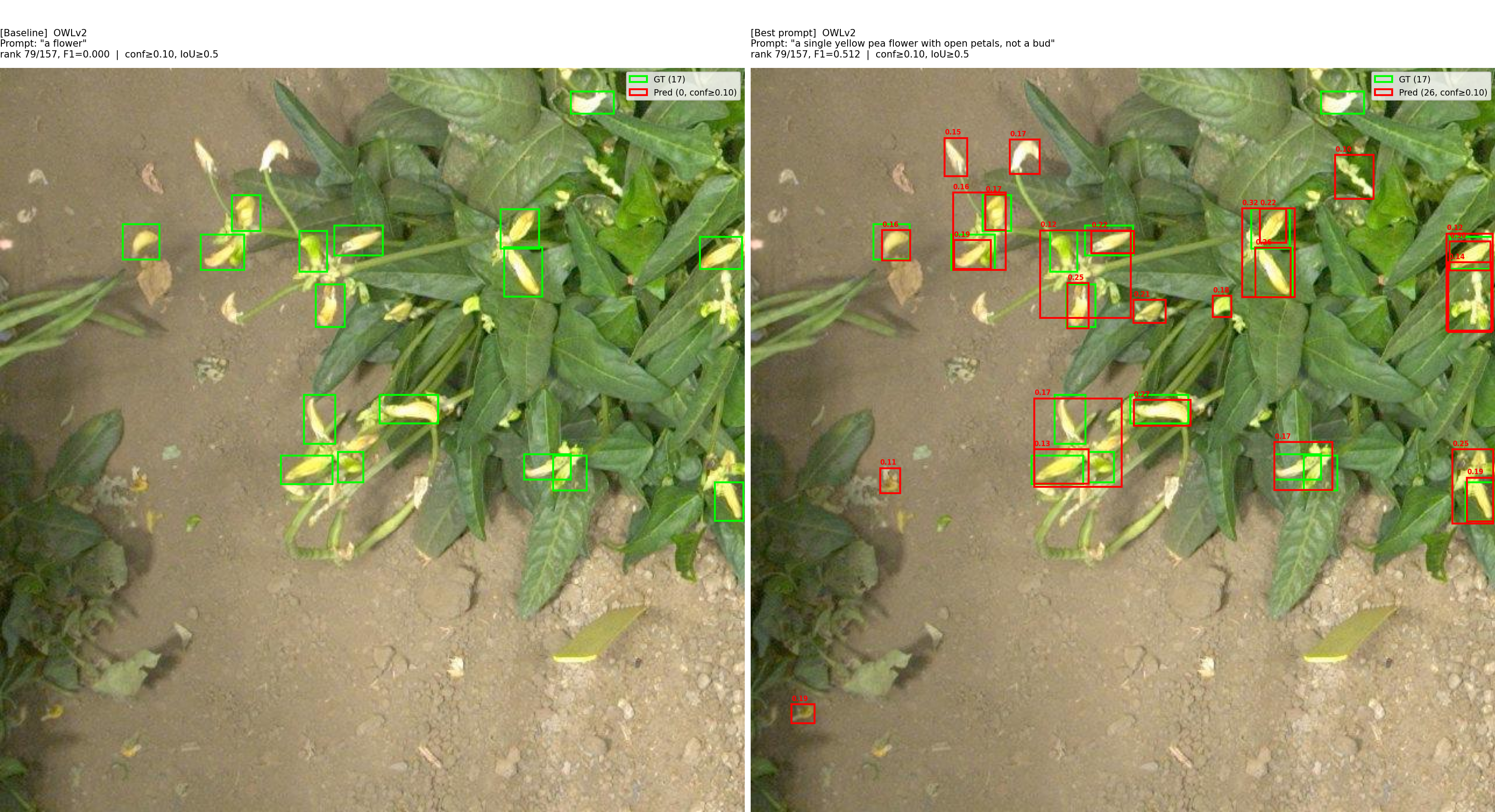} &
        \includegraphics[width=0.47\textwidth]{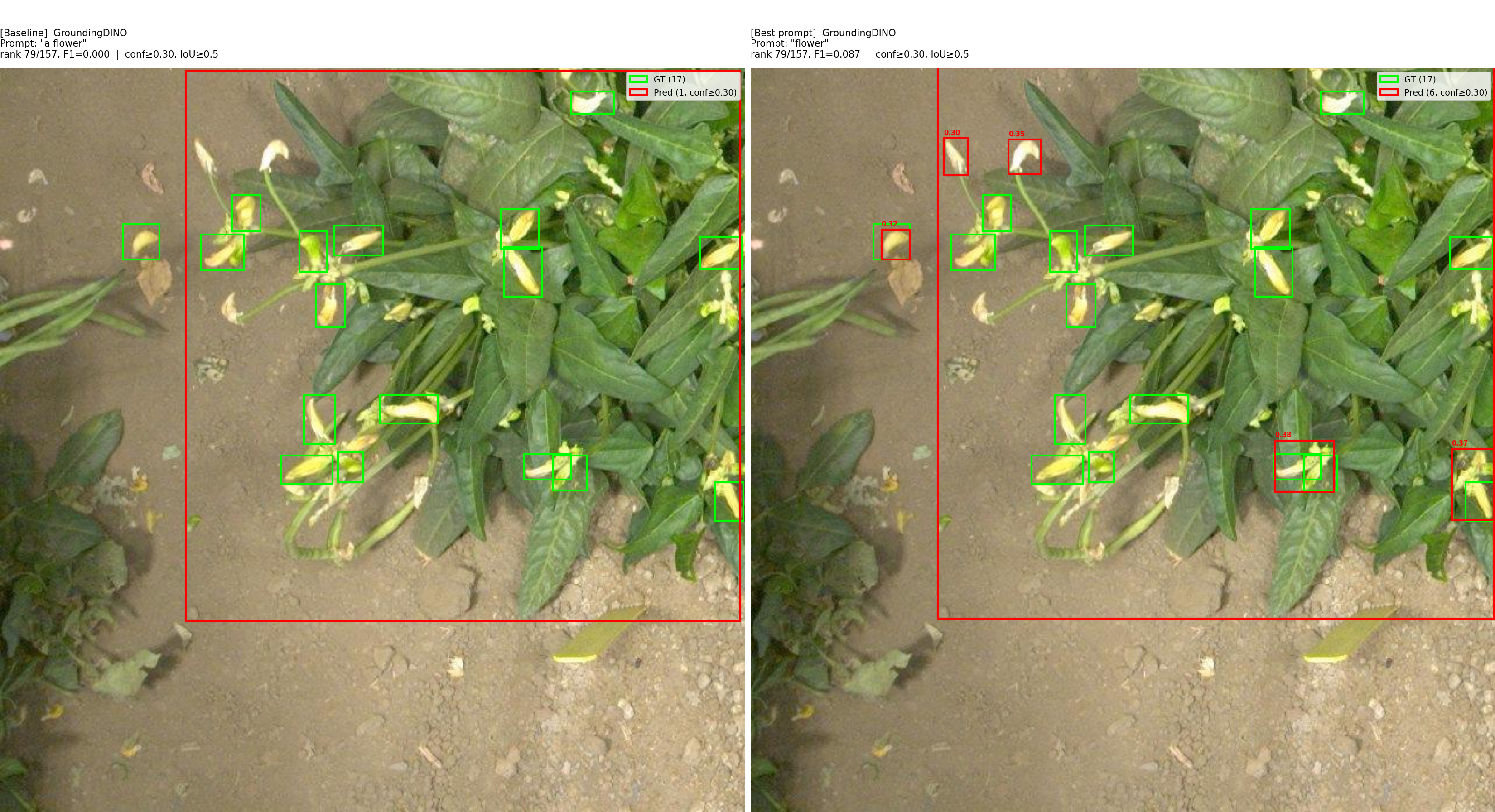} \\
    \end{tabular}
    \caption{Qualitative comparison of baseline (left half:
    \texttt{"a flower"}) versus best discovered usable prompt (right half) across all four
    VFMs on a representative real image (median-ranked by F1).
    Green boxes denote ground-truth annotations; red boxes denote predictions
    above the model's F1-maximizing confidence threshold from synthetic benchmarking.
    YOLO World and OWLv2 have the largest visible performance gain, while Grounding DINO and SAM3 also improve by F1, but less obviously.  }
    \label{fig:qualitative-phase2-real}
\end{figure*}

\section{Discussion}
\label{sec:discussion}
\subsection{Crop and Object Generalization}
The consistent ranking of prompt configurations across synthetic and real
data suggests that the structural preferences discovered---\eg the benefit
of long prompts involving negators and color specificity for YOLO World, or taxonomy for
OWLv2---may reflect how these models encode visual concepts rather than
dataset-specific artifacts. The pod detection experiments
(Section~\ref{sec:exp_pod_generalization}) provide concrete evidence for this
claim: the structural axis preferences discovered on flowers transfer
meaningfully to a morphologically distinct object class on the same crop.
Structured prompts assembled by translating the flower-optimal axis
configurations into pod-appropriate terms yield substantial improvements over
na\"ive baselines for all models, even without running the full optimization
pipeline on pod data. This provides a promising avenue to explore further on other crop and object datasets.

The LLM-guided axis translation further demonstrates that automated prompt
generation can surface descriptors---such as ``mottled'' for SAM3---that a
manual translation from flower axes would be unlikely to produce but provide information that clearly boosts model detection performance. Importantly, the
synthetic-to-real transfer comparison (Table~\ref{tab:syn_real_transfer})
shows that prompts optimized entirely on synthetic data match or exceed
real-data-optimized prompts in the majority of model--object combinations,
for both flowers and pods. This supports the practical viability of the
pipeline for new crops and structures without requiring labeled real-world
data: synthetic benchmarking alone provides a sufficient optimization
substrate. Whether these structural preferences extend beyond cowpea to
other crop species remains an open question; VLM-guided prompting to
automatically identify species, color, and size descriptors for new domains
is one potential direction.

\subsection{Confidence Threshold Sensitivity}
A practical challenge that compounds the prompt sensitivity documented above
is the need to select an appropriate confidence threshold for each
model--prompt combination. Unlike supervised detectors where the confidence
distribution is shaped during training on the target domain, VFMs produce
confidence scores that are uncalibrated with respect to the downstream task:
the same object may receive a score of 0.78 under one prompt and 0.15 under
another, and neither value carries a consistent probabilistic interpretation
across configurations.

This variability is clearly visible in the qualitative results
(Figure~\ref{fig:qualitative-phase2-real}), which shows the median-ranked
image (by F1) for each model. The F1-maximizing confidence thresholds
differ dramatically: SAM3 operates at $\text{conf} \geq 0.41$, Grounding
DINO at $\text{conf} \geq 0.30$, YOLO World at $\text{conf} \geq 0.24$,
and OWLv2 at $\text{conf} \geq 0.10$. A fixed threshold (\eg 0.5) that
would retain most SAM3 detections would suppress nearly all OWLv2 and YOLO
World predictions.

For the qualitative visualizations in Figure~\ref{fig:qualitative-phase2-real},
we apply the confidence threshold that maximizes F1 on the synthetic dataset
to filter predictions on real images. This reflects the practical deployment
scenario in which labeled synthetic data is available for threshold
calibration but the target real domain is unlabelled. The resulting
predictions are therefore subject to the domain gap between synthetic and
real confidence distributions. 

This miscalibration further validates our choice of threshold-free mAP as the optimization metric. mAP integrates over
the full precision--recall curve~\cite{Everingham2010PASCAL, Lin2014COCO},
evaluating the ranking quality of all detections rather than the binary
accept/reject decisions produced by any single threshold. This makes mAP
robust to the cross-model and cross-prompt confidence miscalibration
described above and ensures that comparisons between configurations are not
confounded by threshold selection. 

In practice, a user deploying these models without any labeled reference data
from the target domain must either accept a threshold calibrated on synthetic
or proxy data (as we do for visualization) or collect a small calibration set, negating the annotation-free advantage of this method. The well-documented miscalibration of modern neural networks~\cite{guo2017calibrationmodernneuralnetworks} is compounded in the open-vocabulary setting, where Wang \etal~\cite{wang2024openvocabularycalibrationfinetunedclip} show that existing calibration techniques fail for unseen classes in CLIP-based models. Developing self-calibrating confidence mechanisms for open-vocabulary detectors in domain-shifted agricultural settings remains an open problem.

\section{Conclusion}

In this work, we presented a systematic framework for optimizing text prompts to deploy zero-shot vision foundation models (VFMs) in complex agricultural environments. Our benchmarking across YOLO World, SAM3, Grounding DINO, and OWLv2 reveals that optimal prompting strategies are highly model-specific and often diverge from human-intuitive class descriptions. Through structured one-factor-at-a-time analysis and combinatorial optimization, we demonstrated that carefully engineered prompts developed entirely on synthetic crop imagery transfer effectively to real-world field data, yielding substantial mAP improvements for all but Grounding DINO without requiring manual annotation. The generalization of this framework to cowpea pod detection---a morphologically distinct object on the same crop---validates that the discovered structural axis preferences reflect model-level encoding properties rather than object-specific artifacts. Notably, prompts optimized on synthetic data alone match or exceed those discovered through real-data optimization in the majority of model--object combinations, supporting the practical viability of fully annotation-free prompt engineering. The LLM-guided axis translation further demonstrates that automated prompt generation can surface domain-specific descriptors that manual translation would overlook, as exemplified by the ``mottled'' descriptor that drove large SAM3 gains on pod detection. While uncalibrated confidence scores across models and domains remain a bottleneck for fully autonomous deployment, this study establishes that rigorous, domain-aware prompt engineering is a critical prerequisite for unlocking the zero-shot capabilities of VFMs. Future work should aim to generalize this framework to a larger set of crops and use the model-specific prompting preferences to inform efficient embedding-space optimization.